\documentclass[11pt]{article}

\usepackage[final]{acl}

\usepackage{times}
\usepackage{latexsym}
\usepackage[T1]{fontenc}
\usepackage[utf8]{inputenc}
\usepackage{microtype}
\usepackage{inconsolata}
\usepackage{graphicx}
\usepackage{booktabs}
\usepackage{amsmath}
\usepackage{amssymb}
\usepackage{placeins}
\usepackage{xcolor}

\newcommand{\rev}[1]{#1}

\title{Lexical Coupling in GUI Element Grounding:\\
Sentence Embeddings Track Labels across Mobile and Web}

\author{Qijia Chen \\
  Department of Computer Science \\
  University of Helsinki \\
  Helsinki, Finland \\
  \texttt{qijia.chen@helsinki.fi} \\\And
  Giulio Jacucci \\
  Department of Computer Science \\
  University of Helsinki \\
  Helsinki, Finland \\
  \texttt{giulio.jacucci@helsinki.fi} \\}

\begin{document}
\maketitle

\begin{abstract}
\rev{GUI grounding evaluations that expose UI elements as text metadata
often treat high instruction--element embedding similarity as evidence
of semantic grounding. Across three mobile and web benchmarks, we show
that this interpretation is frequently confounded by visible-label
recovery. Lexical baselines
remain competitive at top-1, label-poor targets remain weak for
text-only methods, and encoder top-1 hits are predictable from lexical
rank, candidate-pool size, and label type.
We evaluate each action as a same-screen ranking task, comparing five
off-the-shelf single-vector encoders with lexical baselines. Encoders
recover some lexical misses, but deployable fusion gains are much smaller
than target-aware oracle gains. These findings show that embedding-based
evaluations can conflate visible-label recovery with semantic GUI grounding.
Embedding-based evaluations should therefore report lexical
baselines, label-type stratification, and deployable-fusion diagnostics.
Our released
repository provides analysis scripts and de-texted per-step panels:}
\url{https://github.com/qijia123/lexical-coupling-release}.
\end{abstract}

\section{Introduction}
\label{sec:intro}

\begin{figure*}[t]
\centering
\includegraphics[width=\linewidth]{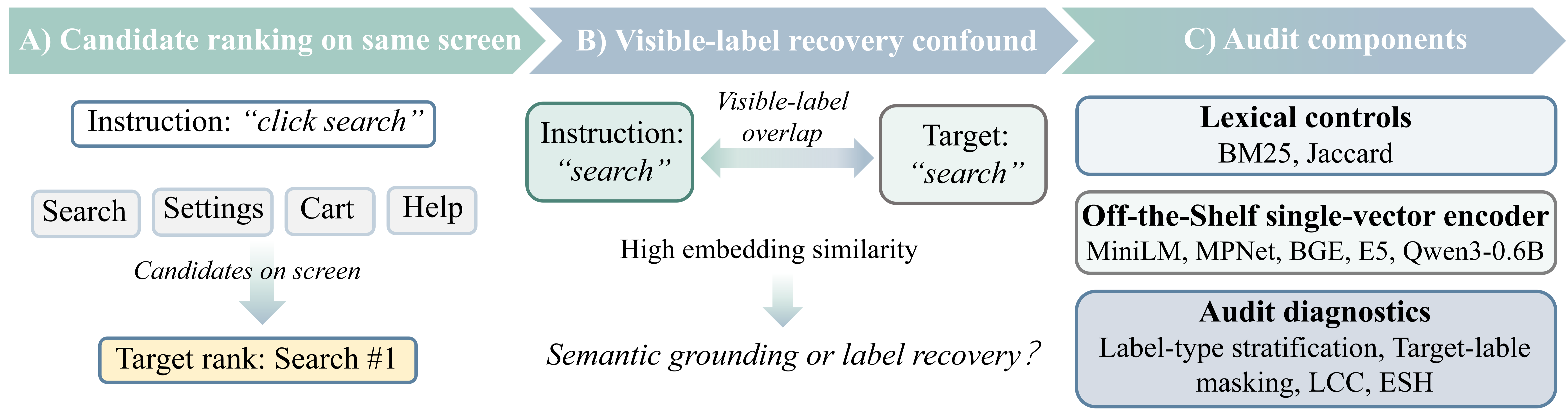}
\caption{\rev{Overview of the measurement audit.
\textbf{(A)} Text-metadata GUI grounding is treated as same-screen
candidate ranking given an instruction.
\textbf{(B)} High instruction--element embedding similarity may reflect
visible-label overlap rather than semantic grounding.
\textbf{(C)} We compare off-the-shelf single-vector encoders against
lexical controls and diagnose the results with label-type stratification,
target-text dependence perturbation, LCC, and ESH.}}
\label{fig:audit-overview}
\end{figure*}

GUI agents must map natural-language instructions to actionable
on-screen elements~\citep{gou2025uground, hong2024cogagent,
cheng2024seeclick, wu2025osatlas}, and benchmarks supply the
element-level targets that make this mapping
measurable~\citep{deng2023mind2web, li2024androidcontrol}. As agents
and benchmarks scale, this mapping is increasingly scored by
instruction--element sentence-embedding similarity over textual
metadata. But high similarity is ambiguous when the instruction repeats
the target's visible label: it may reflect label recovery rather than
semantic inference from task or screen context
(Figure~\ref{fig:audit-overview}).

In Messick's terms~\citep{messick1995validity}, embedding distance is a
valid grounding proxy only if it carries signal beyond a lexical
baseline; this paper is an audit of how GUI grounding is
\emph{measured}, not a proposal for a new grounding model.
\rev{We focus on off-the-shelf single-vector sentence encoders as
plug-and-play similarity metrics, with fine-tuning as a control.} We recast
each grounded step as ranking the visible same-screen UI elements and
compare lexical retrievers with off-the-shelf encoders across
AndroidControl~\citep{li2024androidcontrol},
MoTIF~\citep{burns2022motif}, and Mind2Web~\citep{deng2023mind2web}.
We stratify by target label type, summarise lexical coupling and
fusion headroom with LCC and ESH, and use STS-B~\citep{cer2017semeval}
plus AndroidControl-Curated~\citep{leung2025curated} to bound generic
encoder weakness and annotation noise.

\begin{figure}[t]
\centering
\includegraphics[width=\linewidth]{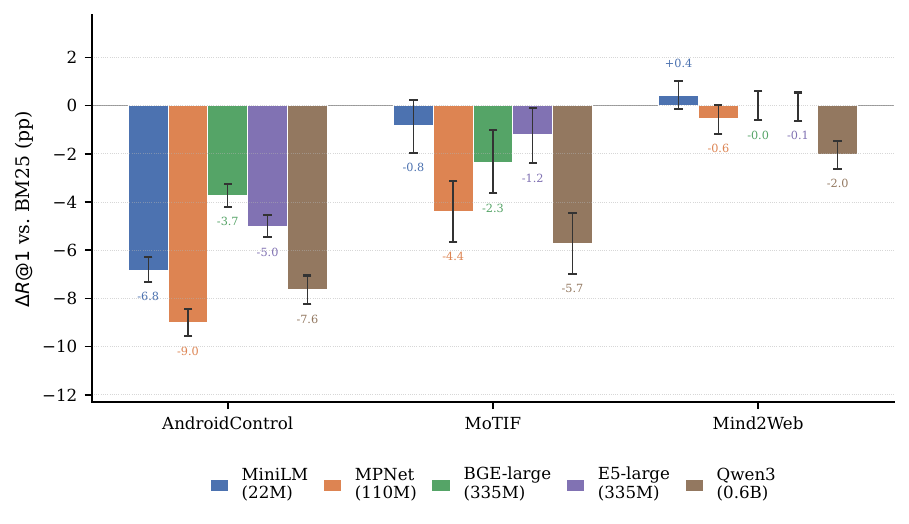}
\caption{Paired $\Delta R@1$ of each sentence-embedding retriever
versus BM25 on the three corpora. Error bars are $95\%$
cluster-bootstrap CIs by episode or \texttt{annotation\_id}.}
\label{fig:cross-corpus}
\end{figure}

\rev{Three findings follow. First, across AndroidControl, MoTIF, and
Mind2Web, none of the five off-the-shelf encoders consistently outperforms
BM25 at $R@1$; Figure~\ref{fig:cross-corpus} shows the paired cross-corpus
deltas. Second, performance depends strongly on exposed target labels:
all text-only methods remain weak on label-poor targets, and
masking the target's exposed text metadata reduces $R@1$ by 22--38
percentage points on an equal-allocation three-stratum sample. Encoder
hit@1 outcomes are also predictable from
lexical rank, label type, and candidate-pool size, with AUCs of
$0.82$--$0.88$. Task-specific fine-tuning substantially improves top-1
ranking accuracy but does not remove lexical coupling: the fine-tuned
model retains an LCC of $0.877$. Its $R@1$ also drops from $0.694$ to
$0.151$ under the target-text dependence perturbation, showing that the
improved ranker continues to rely strongly on text exposed by the target
candidate. Third, embeddings contain complementary
signal, but on AndroidControl, standard deployable fusion captures only a
small fraction of the target-aware oracle headroom and can reduce top-1
accuracy when many correlated retrievers are combined.}

\noindent\textbf{Contributions.}
(1)~\rev{We show that high off-the-shelf instruction--element embedding
similarity often reflects visible-label recovery and therefore does not,
by itself, establish semantic GUI grounding.}
(2)~\rev{We identify a stable cross-corpus mechanism behind this effect:
instructions often name targets using visible UI labels, while benchmark
metadata re-exposes those labels in candidate text.}
(3)~\rev{We develop LCC and ESH and situate them in an audit suite with
lexical baselines, label-type stratification, and a target-text dependence
perturbation,
quantifying lexical coupling and separating target-aware oracle headroom
from deployable fusion gains.}
(4)~\rev{We turn these findings into a practical reporting protocol for
text-metadata GUI grounding evaluations.}

\section{Related Work}
\label{sec:related}

Our audit sits at the intersection of GUI grounding evaluation,
construct-validity and shortcut-learning audits in NLP, and
lexical-versus-dense retrieval in IR.

\paragraph{GUI grounding benchmarks and agents.}
GUI grounding benchmarks evaluate whether models map natural-language
instructions to UI elements or screen coordinates. We use three
text-metadata corpora: AndroidControl~\citep{li2024androidcontrol},
MoTIF~\citep{burns2022motif}, and Mind2Web~\citep{deng2023mind2web}
(sizes and splits in \S\ref{sec:method-data}). Pixel-level visual
grounders~\citep{cheng2024seeclick, wu2025osatlas, gou2025uground, hong2024cogagent}
and end-to-end Android agent harnesses~\citep{xu2025androidlab} lie
outside our text-similarity scope. This literature asks whether
models can ground; we ask what the similarity scores measure.

\paragraph{Construct validity and shortcut learning in NLP.}
We build on construct-validity audits in language understanding.
\citet{messick1995validity} establishes that a score-based metric is
informative only insofar as it carries the construct it claims to
measure. \citet{gururangan2018artifacts} and \citet{mccoy2019hans} show NLI
classifiers can predict labels from annotation artifacts or
lexical-overlap heuristics rather than the entailment relation itself;
\citet{geirhos2020shortcut} frame these as shortcut learning. We adapt
the audit logic but not its verdict: in GUI grounding, label overlap is
often a valid route to the target, so the construct-validity question
is not whether embeddings exploit a shortcut but whether their scores
carry evidence beyond that lexical route. Our lexical baselines,
label-type stratification, and residual logistic audit are that
methodological transfer.

\paragraph{Lexical versus dense retrieval.}
What NLP construct-validity audits call a shortcut, IR treats as a
lexical baseline that should not be skipped: dense and sparse
retrievers must be reported side by side before dense similarity is
interpreted as semantic evidence. BM25~\citep{robertson1994bm25} is the
canonical lexical reference; dense retrieval work such as
DPR~\citep{karpukhin2020dpr} motivates neural semantic matching in
open-domain QA, while BEIR~\citep{thakur2021beir} shows that BM25
remains a robust zero-shot baseline across heterogeneous text-retrieval
tasks. Our setting asks a different measurement question:
when sentence-embedding similarity is used as a GUI grounding metric,
does it carry evidence beyond visible-label recovery? GUI candidate
texts expose the same labels annotators use to describe targets,
making lexical matching both a valid grounding route and a confound
for interpreting embedding similarity as semantic grounding. We turn
this mechanism into reportable lexical coupling and fusion headroom
diagnostics.
Reciprocal-rank fusion~\citep{cormack2009rrf} is a standard
rank-fusion baseline; the encoders we audit are standard MTEB-style
sentence encoders~\citep{muennighoff2023mteb}
(\S\ref{sec:method-retrievers}), and STS-B contrast
(\S\ref{sec:res-sts}) shows the coupling is corpus-specific rather than
evidence of weak encoders.

\paragraph{Audits of GUI grounding quality.}
Two recent audits of GUI grounders sit alongside ours.
\citet{jandial2026guigrounders} use adversarial multi-instruction
generation for a model-side audit of robustness to query variation.
AndroidControl-Curated~\citep{leung2025curated} identifies ambiguities
and factual errors in AndroidControl and releases a purified split,
an annotation-side cleanup that holds the evaluation metric
fixed. Ours is a metric-side audit: we hold the model family
and benchmark fixed and ask what the similarity score itself measures.
We use AndroidControl-Curated's flags in \S\ref{sec:res-curated} as a
robustness check; their benchmark purification and our metric audit
are complementary.

\section{Method}
\label{sec:method}

We operationalise the audit as a lexical-controlled candidate-ranking
protocol that ranks same-screen UI elements with lexical and embedding
retrievers, stratifies results by target label availability, and
summarises lexical coupling and fusion headroom with LCC and ESH.

\subsection{Datasets and candidate pools}
\label{sec:method-data}

\paragraph{AndroidControl.}
AndroidControl~\citep{li2024androidcontrol} provides 15{,}283 mobile
UI demonstration episodes with per-step instructions paired to typed
actions. From the accessibility tree at each grounded step we extract
every visible UI element and resolve the target at the action
coordinate, yielding 58{,}078 grounded steps and 4.56M total
candidates (median 62 per step, max 1{,}657). We use the
official train/val/test split; Appendix~\ref{app:data-prep} gives
extraction and join-key details.

\paragraph{Label-type taxonomy.}
\label{sec:method-labeltype}
For each target element we record which textual affordance is present
in the accessibility tree: text-only (visible
\texttt{text}), cd-only (only
\texttt{content\_description}), both, resource-only
(resource ID only), pseudo (boilerplate strings such as
\texttt{image container} or \texttt{layout header}), or none.

\paragraph{Main and diagnostic subsets.}
Our main analysis uses clean-main: \texttt{click} and
\texttt{long\_press} actions whose target has visible text or content
description ($n=32{,}646$). Label-poor rows
(resource-only/pseudo/none) and the full
58{,}078-step set are reported separately as diagnostics rather than
averaged into the main metric.

\paragraph{MoTIF.}
For cross-corpus stress testing we run the same target-ranking
protocol on MoTIF~\citep{burns2022motif}, using the episode goal as
the query because MoTIF lacks AndroidControl-style step instructions.
After matching preprocessing this yields 22{,}427 grounded steps; the
clean-main subset (text/cd-labelled click targets) contains 4{,}724
rows. The label-type taxonomy and ranking protocol are otherwise
identical.

\paragraph{Mind2Web (cross-domain).}
We run the same protocol on Mind2Web~\citep{deng2023mind2web}'s three
test splits (5{,}941 usable steps after filtering; mean 531, median
410, max 4{,}301 candidates per step) using the released
\texttt{pos+neg\_candidates} pool. The
query is the task plus the prefix of \texttt{action\_reprs} preceding
the target action; we exclude the current step's
\texttt{target\_action\_repr} because it is element-aware and would
leak the target identity. The label-type taxonomy maps to web
counterparts (text-only, aria-only,
both, id/class-only, none); DOM-parsing
details are in Appendix~\ref{app:data-prep}.

\subsection{Auxiliary datasets for control checks}
\label{sec:method-aux}

\paragraph{AndroidControl-Curated.}
AndroidControl-Curated~\citep{leung2025curated} provides
annotation-quality flags that we use to test whether the BM25 lead is
driven by annotation noise; 2{,}946 of its rows join to our clean-main
subset. Per-flag definitions and counts are in
Appendix~\ref{app:data-prep}.

\paragraph{STS-B.}
To contrast corpus-specific lexical coupling against canonical
sentence similarity we evaluate MiniLM, MPNet, BGE, E5, and Qwen3 on the STS-B dev+test
split~\citep{cer2017semeval} (2{,}879 sentence pairs with gold similarity in
$[0,5]$).

\subsection{Candidate-ranking protocol}
\label{sec:method-task}

For each grounded step we form a ranking query $q$ and a candidate
set $C = \{c_1,\dots,c_n\}$ consisting of every parsed UI element on
the current screen. The query is the step instruction
(AndroidControl), the episode goal (MoTIF), or the task plus
preceding action-history prefix (Mind2Web; see
\S\ref{sec:method-data}). The positive candidate is the element at
the recorded target index. A retriever scores each candidate; the
target's rank is its position when candidates are sorted by descending
score, with deterministic tie-break (Appendix~\ref{app:impl}). We
report standard ranking metrics: $R@1$, $R@5$, $R@10$, and mean
reciprocal rank (MRR). All metrics treat each grounded step as one
ranking event and aggregate uniformly across steps within the chosen
subset.

\subsection{Element Textualisation}
\label{sec:method-textual}

\rev{Each UI element is serialised to a single retriever input string
from visible \texttt{text} and accessibility descriptions, with
resource-id or class-name fallbacks when no readable label is present.
Lexical and embedding retrievers consume the same textualised candidate
strings, so any lexical advantage is not due to BM25 receiving richer UI
text. Corpus-specific parsing, tokenisation, and stoplists are in
Appendix~\ref{app:data-prep}; alternative variants used only in
construct correlation are in Appendix~\ref{app:construct}.}

\subsection{Retrievers}
\label{sec:method-retrievers}

\paragraph{Random.}
A reference baseline that ranks candidates uniformly at random per
step. Expected $R@1$ is the per-step mean of $1/n_x$; we report it to
make candidate-pool difficulty explicit.

\paragraph{Jaccard.}
For each candidate we compute the Jaccard coefficient between query
and candidate token sets. Empty sets receive score zero.

\paragraph{BM25.}
We score candidates with BM25 using fixed, standard hyperparameters
($k_1=1.5$, $b=0.75$) with no corpus-specific tuning. Each screen is
treated as its own collection so that document frequencies reflect the
local distractor pool rather than the global vocabulary, matching how a
deployed grounding system would rank elements on a single screen.
Appendix~\ref{app:impl} confirms the per-screen scheme is not what
drives BM25's advantage.

\paragraph{Sentence embeddings.}
We rank candidates with \rev{five off-the-shelf single-vector encoders
spanning 22M to 0.6B parameters:} MiniLM-L6
(\texttt{all-MiniLM-L6-v2}, 22M) and MPNet
(\texttt{all-mpnet-base-v2}, 110M) sentence-transformer
models~\citep{reimers2019sbert}, BGE
(\texttt{BAAI/bge-large-en-v1.5}, 335M)~\citep{xiao2023cpack},
E5 (\texttt{intfloat/e5-large-v2}, 335M), \rev{and Qwen3-Embedding-0.6B
(\texttt{Qwen/Qwen3-Embedding-0.6B}, 0.6B)~\citep{zhang2025qwen3embedding}.}
Embeddings are $\ell_2$-normalised and ranking uses
cosine similarity. Model-specific instructions are respected:
queries to BGE are prefixed with the documented instruction string and
E5 inputs are prefixed with \texttt{query:} and \texttt{passage:} for
the two sides; \rev{Qwen3 queries use the model-card instruction format.}

\subsection{Fusion}
\label{sec:method-fusion}

We test whether the embeddings carry ranking signal beyond BM25 under
fusion, in two ways.

\paragraph{Oracle min-rank ceiling.}
For each step and retriever set $M$, let $r_j$ be the target rank under
retriever $j$. The oracle target rank is
$r^{\mathrm{or}}_M = \min_{j \in M} r_j$. This is a target-aware
per-step selector: it is not deployable, because it uses the true target
to choose the retriever that ranked that target best. We use it as a
headroom diagnostic before testing deployable fusion.

\paragraph{True Reciprocal Rank Fusion.}
For each candidate $c$ and method $j$ with rank $r_j(c)$ we form
\begin{equation}
s_{\mathrm{rrf}}(c) = \sum_{j \in M} \frac{1}{k + r_j(c)},
\quad k = 60,
\end{equation}
following~\citet{cormack2009rrf}. The target's combined rank is
$1 + |\{c \neq t : s_{\mathrm{rrf}}(c) > s_{\mathrm{rrf}}(t)\}|$, with
$s_{\mathrm{rrf}}$ ties resolved by the deterministic tie-break of
\S\ref{sec:method-stats}. Unlike the oracle, this is a deployable
combiner: it uses per-candidate ranks under each method, not the
target's privileged rank. We report both the oracle ceiling and the
true-RRF realisation so that the gap between them quantifies how much
of the oracle headroom is actually captured by a standard
fusion rule.

\rev{We report lexical fusion, single-embedding BM25+encoder fusion
(BGE, E5, Qwen3), and aggregate fusion over all five main encoders; the
full list is given in Table~\ref{tab:fusion}.}

\subsection{LCC and ESH}
\label{sec:method-lcc-esh}

We formalise two scalar diagnostics that summarise an
encoder's behaviour at the (corpus, encoder) cell level. LCC operates
independently of the fusion machinery in \S\ref{sec:method-fusion};
ESH compresses its oracle and RRF combiners into a single headroom
pair.

\paragraph{Definition 1 (Lexical Coupling Coefficient, LCC).}
For a benchmark $\mathcal{D}$ and an encoder $M$, let
$y_M(x) = \mathbb{1}[r_M(x) = 1]$ be the binary indicator that $M$
ranks the true target at position~$1$ on step~$x$. Define the
control vector
\begin{equation}
\begin{aligned}
\phi(x) = \bigl(&\log(1{+}r_{\textsc{bm25}}(x)),\,
                 \log(1{+}r_{\textsc{jaccard}}(x)),\\
               &\log n(x),\,
                 \mathbf{1}_{\ell(x)}\bigr),
\end{aligned}
\end{equation}
where $\mathbf{1}_{\ell(x)}$ is a one-hot encoding of
the target element's label-type partition. The two rank terms are
lexical; we add candidate-pool size $\log n(x)$ and label type as
structural controls so that success a model could obtain from pool
shape or label availability alone is not credited as semantic signal.
The Lexical Coupling Coefficient is the AUC of a logistic
regression of $y_M$ on $\phi$, fit on $\mathcal{D}$ with $\ell_2$
regularisation ($C=1$) on standardised features:
\begin{equation}
\textrm{LCC}(M;\,\mathcal{D}) \;=\; \mathrm{AUC}\!\left(
   \widehat{\Pr}\!\bigl[\,y_M\,\big|\,\phi\,\bigr],\; y_M
\right).
\end{equation}
LCC takes values in $[0.5,\,1.0]$ and measures how reproducible an
encoder's R@1 successes are from these lexical and structural controls
without reference to the encoder itself. At $\textrm{LCC}{=}0.85$ the
controls assign a higher predicted hit probability to a random encoder
hit than to a random miss $85\%$ of the time; we use
$\textrm{LCC} > 0.85$ as a descriptive high-coupling regime, not a
hypothesis-test cutoff, and report the underlying AUCs throughout.

\paragraph{Definition 2 (Effective Semantic Headroom, ESH).}
For an encoder set $\mathcal{S} = \{M_1,\dots,M_k\}$ and the oracle
min-rank combiner of \S\ref{sec:method-fusion},
\begin{align}
\textrm{ESH}^{\textrm{oracle}}(\mathcal{S};\,\mathcal{D})
   &= \Pr\!\bigl[
        r_{\mathcal{S} \cup \{\textsc{bm25}\}}^{\textrm{min}}(x) = 1
      \bigr] \notag\\
   &\quad - \Pr\!\bigl[r_{\textsc{bm25}}(x) = 1\bigr].
\end{align}
It measures the R@1 gain of the target-aware per-step selector when
$\mathcal{S}$ is added to BM25, and is therefore a headroom diagnostic,
not a deployable fusion result. Its deployable counterpart,
$\textrm{ESH}^{\textrm{real}}(\mathcal{S};\,\mathcal{D})$, is defined
identically with the RRF combiner of \S\ref{sec:method-fusion}
($k=60$~\citep{cormack2009rrf}) replacing the oracle, and is the R@1
gain a standard hybrid retriever actually delivers. The pair
$(\textrm{ESH}^{\textrm{oracle}}, \textrm{ESH}^{\textrm{real}})$
summarises both the in-principle complementarity of $\mathcal{S}$ and
what a conventional fusion rule realises.

Matching thresholds for unrealisable headroom
($\textrm{ESH}^{\textrm{oracle}} - \textrm{ESH}^{\textrm{real}}$) and
destructive fusion ($\textrm{ESH}^{\textrm{real}} \leq 0$), together
with per-retriever rank caching requirements, are in
Appendix~\ref{app:lcc-thresholds}.

\subsection{Fine-tuning control}
\label{sec:method-finetune}

\rev{We include a task-supervised positive control to test whether the
observed limitations are specific to plug-and-play similarity metrics
rather than dense-retriever capacity under task supervision.
MiniLM-L6-v2 (22M) is fine-tuned for one contrastive epoch on the
AndroidControl train split and evaluated on the held-out clean-main test
subset under the same ranking protocol. We then rerun the LCC audit and
target-text dependence perturbation on the fine-tuned ranks to measure
lexical coupling and reliance on target-exposed text after supervision.
Results are in
\S\ref{sec:res-finetune}; full training details are in
Appendix~\ref{app:finetune}.}

\subsection{Statistical inference and reproducibility}
\label{sec:method-stats}

All paired comparisons use a paired cluster bootstrap with $B=2000$
resamples by corpus-specific unit: \texttt{episode\_id} for
AndroidControl and MoTIF, and \texttt{annotation\_id} for Mind2Web.
95\% CIs are percentile intervals.
Ranking ties are broken with a hash-based deterministic jitter keyed
on (episode id, step index, retriever, candidate index) so rank
assignment is reproducible across subsets. Appendix~\ref{app:impl}
gives full inference and implementation details.
\rev{Textualisation code, ranking scripts, and de-texted per-step panels
are released alongside the paper (Appendix~\ref{app:artifacts}).}

\section{Results}
\label{sec:results}

Single-cell numbers are verified against per-step CSV panels released
with the paper; all paired comparisons use a $B{=}2000$ paired cluster
bootstrap on the corpus-specific unit defined in
\S\ref{sec:method-stats}.

\subsection{Target ranking and label availability on AndroidControl}
\label{sec:res-ranking}

On the clean-main subset ($n{=}32{,}646$ steps over $12{,}393$
episodes; \S\ref{sec:method-data}), with random ranking giving
$R@1{=}0.022$ as the difficulty floor, BM25 obtains
$R@1{=}0.557$ and $\text{MRR}{=}0.632$, exceeding every individual
sentence-embedding baseline on top-1 accuracy and MRR
(Table~\ref{tab:q7-main}). \rev{The three larger encoders in the main
comparison, BGE-large, E5-large-v2, and Qwen3-Embedding-0.6B, underperform
BM25 by $3.7$, $5.0$, and $7.6$~pp on $R@1$, respectively; a more recent
0.6B single-vector embedding control does not alter the top-1 pattern.}
Embeddings recover advantage only at $R@5$ and beyond: BGE-large
attains the best $R@5{=}0.724$ and $R@10{=}0.785$, a pattern
consistent with embeddings finding the target in a broader top-$k$
shortlist but ranking it less reliably at position~1. Paired
cluster-bootstrap deltas confirm the BM25 lead over all \rev{five}
encoders on both $R@1$ and MRR; the smallest is BGE-large at $+3.7$~pp
$R@1$ and $+1.6$~pp MRR, with all CIs excluding zero
(Appendix~\ref{app:deltas}).

\begin{table}[t]
\centering\small
\begin{tabular}{l cccc}
\toprule
Retriever  & $R@1$ & $R@5$ & $R@10$ & MRR \\
\midrule
Random     & 0.022 & 0.107 & 0.212 & 0.089 \\
Jaccard    & 0.547 & 0.703 & 0.741 & 0.624 \\
\textbf{BM25} & \textbf{0.557} & 0.707 & 0.742 & \textbf{0.632} \\
MiniLM     & 0.489 & 0.709 & 0.768 & 0.592 \\
MPNet      & 0.467 & 0.716 & 0.783 & 0.583 \\
BGE-large  & 0.520 & \textbf{0.724} & \textbf{0.785} & 0.616 \\
\rev{E5-large}  & \rev{0.507} & \rev{0.707} & \rev{0.767} & \rev{0.603} \\
\rev{Qwen3-Emb.} & \rev{0.481} & \rev{0.680} & \rev{0.740} & \rev{0.577} \\
\bottomrule
\end{tabular}
\caption{Target-ranking metrics on the clean-main subset
($n{=}32{,}646$).}
\label{tab:q7-main}
\end{table}

\paragraph{Mechanism: label-type stratification.}
Stratifying by target label type
(Table~\ref{tab:q7-labeltype}; $n{=}58{,}078$ all DB rows) separates
three regimes. On text-rich targets, lexical and embedding
methods both work but BM25 wins top-1; on
content-description-only targets all methods sit within a few
points of each other; on label-poor targets all methods remain weak.
The 25{,}430 resource-only/pseudo/none rows show the boundary: pseudo
and none rows sit at random scale, while resource-only rows top out at
\rev{$R@1{=}0.124$ under Qwen3}, far below text-rich performance. Thus
embedding alignment does not reliably recover targets that lack a
human-readable label; the same low-label collapse holds across corpora
(Appendix~\ref{app:label-collapse}).
Masking the target's exposed text metadata drops $R@1$ by $22$--$38$~pp
on an equal-allocation stratified sample of text-only, cd-only, and
both-labelled targets (Appendix~\ref{app:label-mask}), providing a
controlled measure of target-text dependence that complements the
cross-corpus label-type collapse.
The \texttt{input\_text} diagnostic subset, excluded from clean-main
because the DB target resolves to the system keyboard rather than the
intended input field, is documented in Appendix~\ref{app:data-prep}.

\begin{table}[t]
\centering\scriptsize
\setlength{\tabcolsep}{1.2pt}
\begin{tabular}{@{}l r cccccccc@{}}
\toprule
type & $n$ & Rand. & Jacc. & BM25 & Mini & MPNet & BGE & \rev{E5} & \rev{Qwen3} \\
\midrule
text-only     & 21{,}539 & 0.023 & 0.612 & \textbf{0.624} & 0.520 & 0.504 & 0.563 & \rev{0.552} & \rev{0.507} \\
both           &  2{,}668 & 0.016 & 0.531 & 0.530 & 0.548 & 0.398 & \textbf{0.549} & \rev{0.484} & \rev{0.504} \\
cd-only       &  8{,}441 & 0.023 & 0.386 & 0.396 & 0.392 & 0.397 & 0.402 & \rev{0.401} & \rev{\textbf{0.407}} \\
resource-only &  9{,}669 & 0.018 & 0.063 & 0.064 & 0.085 & 0.085 & 0.099 & \rev{0.095} & \rev{\textbf{0.124}} \\
pseudo         &  2{,}629 & 0.009 & 0.010 & \textbf{0.011} & 0.008 & 0.008 & 0.007 & \rev{0.006} & \rev{0.005} \\
none           & 13{,}132 & 0.027 & 0.013 & 0.013 & 0.016 & 0.011 & 0.006 & \rev{0.006} & \rev{\textbf{0.020}} \\
\bottomrule
\end{tabular}
\caption{$R@1$ by target label-type on all DB rows.
Best non-random per row in bold.}
\label{tab:q7-labeltype}
\end{table}

\rev{To complement the quantitative label-type and masking analyses,
Appendix~\ref{app:diagnostic-examples} provides a qualitative error
analysis with de-identified examples for the three recurring regimes:
lexical label recovery, residual embedding recovery, and label-poor
failure. These examples are illustrative; the population-level evidence is
Table~\ref{tab:q7-labeltype} and the target-text dependence perturbation in
Appendix~\ref{app:label-mask}.}

\subsection{Cross-corpus generalisation on MoTIF and Mind2Web}
\label{sec:res-replication}

Because the corpora differ simultaneously in query granularity, domain,
candidate-pool size, and annotation protocol, this section tests
whether the within-corpus BM25-vs-embedding gap \emph{generalises}
across these conditions. We therefore interpret cross-corpus absolute
scores descriptively and focus on the paired $\Delta$ within each
corpus, where BM25 and the encoder see identical queries and candidate
pools.

We apply the same candidate-ranking audit to MoTIF (mobile, episode
goal as query) and Mind2Web (web, task-with-history;
\S\ref{sec:method-data}): on neither corpus does an off-the-shelf
embedding significantly beat BM25 at top-1, and the low-label weak regime
recurs on both, across independent corpora collected
under different annotation protocols (Table~\ref{tab:replication}). On 4{,}724 MoTIF
clean-main click targets, BM25 wins top-1 with $R@1{=}0.224$ versus
\rev{E5-large's $R@1{=}0.212$,} BGE-large's $R@1{=}0.201$,
\rev{and Qwen3-Embedding-0.6B's $R@1{=}0.167$;} absolute scores are lower than on
AndroidControl because the query is a goal rather than a step
instruction.
Embeddings recover advantage further down the list: MiniLM attains
the best MRR ($0.342$), BGE-large the best $R@5$ ($0.479$), and MPNet
the best $R@10$ ($0.633$). On
the union of the three Mind2Web test splits ($n{=}5{,}941$ usable
steps; pool median $410$), no encoder significantly exceeds BM25 at
$R@1$ (BGE-large and \rev{E5-large} tie within paired CIs, while
\rev{Qwen3 is below BM25}); BM25
retains significant MRR leads over MPNet, BGE-large, \rev{E5-large, and Qwen3}. The low-label
weak regime recurs on both: on MoTIF's 17{,}703 label-poor rows
$R@1$ ranges only $0.039$--$0.073$ against random $0.042$, and on
Mind2Web's 474 \texttt{id}/\texttt{class}-only or unlabelled rows every
retriever scores $R@1 \leq 0.005$.

\begin{table*}[t]
\centering\small
\begin{tabular}{l cccc cccc}
\toprule
& \multicolumn{4}{c}{MoTIF ($n{=}4{,}724$)} & \multicolumn{4}{c}{Mind2Web ($n{=}5{,}941$)} \\
\cmidrule(lr){2-5}\cmidrule(lr){6-9}
Retriever  & $R@1$ & $R@5$ & $R@10$ & MRR & $R@1$ & $R@5$ & $R@10$ & MRR \\
\midrule
Random        & 0.022 & 0.131 & 0.262 & 0.101 & 0.005 & 0.020 & 0.036 & 0.022 \\
Jaccard       & 0.217 & 0.386 & 0.483 & 0.314 & 0.039 & 0.115 & 0.177 & 0.088 \\
\textbf{BM25} & \textbf{0.224} & 0.391 & 0.503 & 0.320 & 0.041 & \textbf{0.125} & \textbf{0.195} & \textbf{0.094} \\
MiniLM        & 0.216 & 0.458 & 0.631 & \textbf{0.342} & 0.045 & 0.116 & 0.165 & 0.090 \\
MPNet         & 0.180 & 0.459 & \textbf{0.633} & 0.316 & 0.035 & 0.091 & 0.135 & 0.073 \\
BGE-large     & 0.201 & \textbf{0.479} & 0.627 & 0.338 & 0.040 & 0.106 & 0.151 & 0.084 \\
\rev{E5-large} & \rev{0.212} & \rev{0.444} & \rev{0.579} & \rev{0.328} & \rev{0.040} & \rev{0.114} & \rev{0.164} & \rev{0.087} \\
\rev{Qwen3-Emb.} & \rev{0.167} & \rev{0.392} & \rev{0.552} & \rev{0.288} & \rev{0.020} & \rev{0.063} & \rev{0.100} & \rev{0.055} \\
\bottomrule
\end{tabular}
\caption{Cross-corpus target-ranking metrics on MoTIF and Mind2Web.
Bold marks the best non-random value in each metric column.}
\label{tab:replication}
\end{table*}

\subsection{Corpus-invariant lexical coupling and bounded headroom}
\label{sec:res-cross}

Lexical coupling is stable across corpora, while oracle semantic
headroom remains bounded; the LCC and oracle ESH diagnostics
(\S\ref{sec:method-lcc-esh}) quantify both at the encoder level
(Table~\ref{tab:cross-corpus}).
\rev{BGE-large, E5-large, and Qwen3-Embedding-0.6B remain strongly coupled
across corpora (BGE LCC $0.845$--$0.880$; E5 LCC $0.849$--$0.882$;
Qwen3 LCC $0.835$--$0.860$):} an
encoder's R@1 successes are predictable from
$(\log r_{\text{BM25}}, \log r_{\text{Jaccard}}, \log n_{\text{cands}},
\text{label-type})$ on every corpus we test.
Oracle ESH(BM25\,$+$\,BGE) ranges from $+0.027$ to $+0.071$ across
corpora; Mind2Web has the smallest oracle headroom because its
larger and more diverse candidate pool diffuses min-rank gains.

\begin{table}[t]
\centering\scriptsize
\setlength{\tabcolsep}{1.2pt}
\begin{tabular}{@{}l r c c c c c c c@{}}
\toprule
& & & \multicolumn{5}{c}{LCC} & \\
\cmidrule(lr){4-8}
Corpus & $n$ & BM25 & BGE & \rev{E5} & \rev{Qwen3} & MPNet & Mini & ESH$_{\mathrm{or}}$ \\
\midrule
AC main  & 32,646 & 0.557 & 0.880 & \rev{\textbf{0.882}} & \rev{0.860} & 0.834 & 0.856 & $+0.070$ \\
MoTIF    &  4,724 & 0.224 & 0.845 & \rev{\textbf{0.874}} & \rev{0.851} & 0.825 & 0.868 & $+0.064$ \\
M2W all  &  5,941 & 0.041 & \textbf{0.867} & \rev{0.861} & \rev{0.845} & 0.844 & 0.849 & $+0.027$ \\
M2W rich &  5,467 & 0.044 & \textbf{0.855} & \rev{0.849} & \rev{0.835} & 0.839 & 0.836 & $+0.029$ \\
\bottomrule
\end{tabular}
\caption{Cross-corpus LCC and oracle ESH. BM25 is $R@1$;
ESH$_{\mathrm{or}}$ is $\textrm{ESH}^{\textrm{oracle}}$(BM25$+$BGE).}
\label{tab:cross-corpus}
\end{table}

\subsection{Robustness and fine-tuning control}
\label{sec:res-robust}

\paragraph{Fine-tuning control.}
\label{sec:res-finetune}
\rev{As a task-supervised positive control, we fine-tune MiniLM-L6-v2 for
one contrastive epoch on AndroidControl and evaluate it on the held-out
$2{,}949$-row clean-main test subset under the same ranking protocol.}
The fine-tuned encoder reaches $R@1{=}0.694$ overall ($+13.6$~pp over
BM25, $+17.6$~pp over zero-shot MiniLM) and recovers $37.4\%$ of
BM25-miss rows at rank~1. \rev{However, its hits remain predictable from
lexical controls (LCC $0.877$), so task supervision improves the ranker
without removing the lexical-coupling pattern. Its $R@1$ also drops from
$0.694$ to $0.151$ under the target-text dependence perturbation
(Appendix~\ref{app:finetune},
Tables~\ref{tab:finetune-coupling}--\ref{tab:finetune-mask}), showing that
the supervised gains remain strongly dependent on target-exposed text.}

\rev{\paragraph{Ruling out annotation noise and generic encoder weakness.}%
\label{sec:res-curated}\label{sec:res-sts}%
Two controls bound this interpretation: BM25 still leads on reviewed
AndroidControl-Curated rows, while all five encoders strongly
outperform Jaccard on STS-B. Thus the pattern is not explained by annotation
noise or generic encoder weakness; full counts and STS-B breakdowns are in
Appendices~\ref{app:data-prep} and~\ref{app:sts}.}

\subsection{Residual signal and deployable fusion}
\label{sec:res-residual}

\paragraph{Conditional recovery.}
Embeddings recover $14$--$16\%$ of clean-main BM25-miss rows at
rank~1, concentrated on label-rich targets (BGE recovers $23.3\%$ of
both-labelled misses); they agree with BM25 on $80.9\%$ of
BM25-hit rows: redundant on easy cases, disagreeing only on a
non-trivial residual.

\paragraph{Oracle vs.\ deployable fusion.}
\rev{The min-rank oracle over BM25, Jaccard, and all five main encoders
attains $R@1{=}0.706$ on clean-main ($+0.151$ over BM25). The lift is
concentrated where labels expose multiple affordances: $+0.227$ on both,
$+0.210$ on cd-only, and $+0.119$ on text-only. True RRF over the same
candidate-level ranks leaves most oracle headroom unrealised
(Appendix~\ref{app:fusion}, Figure~\ref{fig:oracle-vs-rrf}): single-embedding
pairs BM25\,$+$\,BGE, BM25\,$+$\,E5, and BM25\,$+$\,Qwen3 gain only
$+0.012$, $+0.011$, and $+0.005$ $R@1$, respectively, and the
lexical-plus-five-embedding combiner whose oracle lifts by $+0.151$
instead \emph{decreases} $R@1$ by $0.009$ under deployable RRF. The
largest deployable top-1 fusion gain remains the lexical pair
BM25\,$+$\,Jaccard ($R@1{=}0.578$,
$+0.023$~[$+0.021,+0.025$] over BM25); the full oracle-vs-RRF table
is in Appendix~\ref{app:fusion}.} Oracle headroom identifies
complementary signal but substantially overstates deployable fusion gains;
\rev{validation-tuned score fusion reaches the same conclusion
(Appendix~\ref{app:calibrated-fusion}).}

\section{Discussion}
\label{sec:discussion}

\paragraph{Two theses.}
The results support two claims. Off-the-shelf sentence-embedding
similarity is not a sufficient grounding proxy without lexical controls.
Its top-1 behaviour is largely predictable from lexical rank and
candidate structure, and \rev{deployable RRF gains remain far smaller
than target-aware oracle headroom.
Fine-tuning sharpens this conclusion. It substantially improves ranking,
ruling out a failure of dense retrieval as a class, but the resulting
model remains lexically coupled and label-sensitive.}

\paragraph{Similarity alone is not grounding evidence.}
Sentence embeddings contribute real signal to GUI grounding, but on
their own their similarity does not establish semantic grounding
without lexical controls. Across the corpora, instruction--label overlap
explains much of top-1 target recovery because users name elements by
visible words that UI metadata re-exposes as text or accessibility
labels. $\textrm{LCC}(\text{BGE-large})$
remains in $[0.85, 0.88]$ across mobile and web
(Table~\ref{tab:cross-corpus}), so this is a property of GUI candidate
spaces and not of any one corpus. In that setting, a high embedding
score can mean that a model recovered the visible label, not that it
inferred the target from task semantics, screen context, or
affordance. Removing only the target candidate's visible label, while
leaving non-target labels intact, drops $R@1$ by $22$--$38$~pp across
BM25 and the embeddings on the same equal-allocation stratified subset
(Appendix~\ref{app:label-mask}). This controlled drop shows that exposed
target text supports a substantial share of top-1 recovery. Together
with the lexical baselines, label-type stratification, and LCC results,
it establishes visible-label recovery as a major component of performance
in this evaluation setting. This matters because many GUI-agent evaluations
use alignment scores as evidence that an agent understands an
instruction. Our results show that such a construct-validity claim
requires evidence beyond embedding distance alone.

\paragraph{Embeddings still carry residual signal.}
Embeddings recover $\sim 15\%$ of
BM25-miss rows and improve deeper-recall coverage, but standard RRF
captures only a small part of this residual and aggregate combiners can
lose top-1 accuracy, a pattern we name destructive fusion
(\S\ref{sec:method-lcc-esh}). Highly correlated, lexically coupled
retrievers may aggregate without adding enough target information,
while their disagreements on non-target candidates can accumulate. The gap between oracle and
deployable headroom is itself useful: it separates the existence of
residual non-lexical signal from the claim that a practical system can
exploit it.

\paragraph{Two deployment paths.}
For deployed GUI agents, off-the-shelf embeddings should not be the
sole top-1 ranker. Two paths follow: lexical retrieval followed by
semantic or multimodal reranking, and task-specific fine-tuning, where
even a small encoder escapes the off-the-shelf ceiling on AndroidControl
(\S\ref{sec:res-finetune}). The same lexical-control argument extends to
multimodal GUI evaluations: improvements on text-label-heavy benchmarks
should be compared against text-only lexical retrieval before they
count as visual grounding.

\paragraph{\rev{Reporting recommendations.}}
\rev{Among the controls we examine, label-type stratification is the most
revealing: aggregate scores merge two qualitatively different regimes.
When targets expose readable labels, lexical retrieval is already strong;
when they do not, both lexical and text-only embedding methods perform
poorly. We therefore recommend that text-metadata GUI grounding evaluations
report (1) lexical baselines using the same candidate pool and element
textualization; (2) performance stratified by target label type; (3) results
by candidate-pool-size bucket; (4) target-text dependence perturbations
where feasible; and
(5) deployable fusion gains separately from target-aware oracle headroom.
Without these controls, aggregate scores can be dominated by label-explicit
cases and obscure failures where the exposed candidate text is insufficient.}

\paragraph{\rev{Benchmark-design recommendations.}}
\rev{Lexical coupling should also inform future corpus construction.
Demonstration-style annotation can encourage label-explicit
instructions: for example, an annotator may write ``tap Settings'' while
viewing a button labeled ``Settings.'' Future benchmarks should therefore
(1) annotate how directly each instruction mentions or paraphrases the
target's exposed label; (2) balance label-explicit examples with cases in
which identifying the target requires broader task or screen context,
including visual evidence where relevant; and (3) where feasible, collect
paired instructions for the same screen and target, one with and one
without a direct label mention. Such native pairs complement post-hoc
masking by separating direct label recovery from grounding that relies on
broader contextual evidence.}

\section*{Limitations}

\rev{Our protocol targets text-metadata candidate-ranking evaluations with
enumerable UI-element candidate pools (AndroidControl, MoTIF,
Mind2Web), where elements are represented by accessibility-tree or DOM
text; it does not cover screenshot-, OCR-, or layout-only visual grounding
without additional perception controls.} Image-and-bounding-box benchmarks such as
ScreenSpot~\citep{cheng2024seeclick} and its OS-Atlas
refinement~\citep{wu2025osatlas} require OCR or UI parsing, which
introduces a separate perception variable.

Our claim is bounded to grounding-metric validity rather than downstream
behavioural-difficulty prediction: step-level embedding distances do
not predict local repair actions on AndroidControl, where repair
signals are sparse by design (Appendix~\ref{app:validity}).

Our conclusions concern off-the-shelf single-vector sentence encoders
under candidate-ranking evaluation. \rev{Reciprocal-rank fusion and a
validation-tuned linear score-fusion stress test leave most oracle headroom
unrealised (Appendix~\ref{app:calibrated-fusion}); richer learned combiners
remain future work.}

One epoch of contrastive fine-tuning substantially improves a 22M
encoder on the same protocol (\S\ref{sec:res-finetune}); stronger
visual representations or end-to-end agent training may change
absolute performance. The measurement recommendation remains: read
such gains relative to lexical controls and a label-type breakdown.

\bibliography{paper}

\appendix

\section*{Appendix overview}
The appendix is organised as follows. Appendices~\ref{app:artifacts}--\ref{app:impl}
cover reproducibility: artifact licensing and the data statement, data
preprocessing per corpus, LCC/ESH thresholds and rank-caching, and full
statistical-inference and implementation details. Appendices~\ref{app:deltas}--\ref{app:finetune}
and~\ref{app:sts} provide supplementary tables and analyses underlying the
main results, including paired BM25-vs-encoder deltas, label-type heatmaps,
the target-text dependence perturbation, LCC cross-validated AUC, oracle
min-rank vs.\ RRF,
calibrated fusion, fine-tuning breakdowns, and STS-B per-encoder Spearman. Appendices~\ref{app:construct}
and~\ref{app:validity} document the construct-validity boundary: the
$M_3$-to-Jaccard convergence and the behavioural validity limit our claims
respect. Appendix~\ref{app:diagnostic-examples} provides a qualitative
error analysis with de-identified examples for the three ranking regimes
discussed in \S\ref{sec:results}; Appendix~\ref{app:stress-tests} adds
Qwen3 paired intervals and the sparse-retrieval stress test.

\section{Artifact use, license, and data statement}
\label{app:artifacts}

We use five public research datasets (AndroidControl, MoTIF, Mind2Web,
AndroidControl-Curated, and STS-B), five public sentence-embedding
checkpoints (MiniLM, MPNet, BGE, E5, and Qwen3-Embedding-0.6B), and one
public sparse-retrieval checkpoint (SPLADE). We cite their creators in
Sections~\ref{sec:related} and~\ref{sec:method} and
Appendix~\ref{app:stress-tests}, and use each only for the evaluation purpose
it was released for.

We do not redistribute the original datasets or checkpoints. We release
only derived per-step evaluation panels (join keys and
numeric/categorical fields), analysis scripts, and summary reports,
under the MIT License and subject to the terms of the original
artifacts; to recover the raw inputs, users fetch the corpora and
checkpoints from their official sources. The code and panels are available in the
project repository linked in the abstract.

Some source screens carry third-party content; AndroidControl Gmail
screens, for instance, can expose user email addresses. We therefore
strip all free-text columns (instructions, element descriptions, DOM
text, STS sentence pairs) from the released panels, keeping only join
keys and numeric/categorical fields, and every numeric result in the
paper is reproducible from these de-texted panels. We do not otherwise
screen the corpora for offensive content beyond their providers'
curation.

\section{Element textualisation and per-corpus preprocessing}
\label{app:data-prep}

\paragraph{AndroidControl.}
The corpus provides 83{,}848 actions across 8 typed action classes
(\texttt{click}, \texttt{long\_press}, \texttt{input\_text},
\texttt{scroll}, \texttt{wait}, \texttt{navigate\_back},
\texttt{navigate\_home}, \texttt{open\_app}). We read all 20 TFRecord
shards under the official 13{,}603/137/1{,}543-episode train/val/test
split. For each tap/long-press/text step we parse the accessibility
tree of the on-screen frame and enumerate every visible UI element
with non-empty bounds. The target element is the smallest-area
visible element whose bounding box contains the recorded action
coordinate. When multiple boxes coincide, we tie-break deterministically
on tree-traversal index. The join key
$(\texttt{episode\_id},\texttt{step\_index})$ is exact and verified
between the action trace and the accessibility-tree screen extracts.

\paragraph{Diagnostic subsets.}
The \texttt{input\_text} diagnostic subset ($n{=}6{,}033$; excluded
from clean-main) is included as an illustration of target-resolution
failure: DB target resolution for text-entry actions lands on the
system keyboard rather than the input field, so every retriever
scores it as if it were a different target. All retrievers fall
\emph{below} random on this subset: random $R@1{=}0.034$, BM25
$R@1{=}0.021$, BGE-large $R@1{=}0.004$. The
clean-no-dup variant ($n=26{,}832$) removes within-episode
duplicate step instructions, which arise when annotators recycle the
previous step's instruction for a \texttt{wait} action and would
inflate spurious M3--friction correlations because the recycled
instruction no longer describes the visible target.

\paragraph{Mind2Web DOM parsing.}
For each candidate identified by \texttt{backend\_node\_id} in the
released \texttt{cleaned\_html}, we aggregate descendant
\texttt{<text>} content, \texttt{aria-label}, \texttt{alt},
\texttt{placeholder}, and \texttt{title} attributes into a textual
representation. The label-type taxonomy maps as
text-only = visible inner text with no aria-label;
aria-only = aria-label / title / placeholder only;
both = both; id/class-only = only DOM \texttt{id}
or \texttt{class}; none = tag only. After filtering steps
with zero or multiple positive candidates the usable set contains
5{,}941 rows (\texttt{test\_domain} 3{,}780, \texttt{test\_task}
1{,}225, \texttt{test\_website} 936); the label-rich subset
(text- or aria-labelled targets, used as ``M2W rich'' in
Table~\ref{tab:cross-corpus}) contains 5{,}467 rows.

\paragraph{AndroidControl-Curated flags.}
\rev{We use the released AndroidControl-Curated quality flags to test
whether the BM25 lead is driven by annotation noise. Of 8{,}377
Curated rows, 2{,}946 fall in our clean-main subset; the relevant
flags cover task rewrites (527 full; 144 clean-main),
candidate-disagreement review (1{,}272 full; 274 clean-main), and
action-text typo fixes (86 full; 0 clean-main). Paired $R@1$ deltas
of BM25 over BGE-large on these rows (cluster bootstrap by episode,
$B{=}2000$) are: overall $+4.4$~pp [$+2.9,+6.0$],
candidate-disagreement review $+8.8$~pp [$+3.3,+14.3$], and task
rewrites $+6.3$~pp [$+0.0,+12.9$] (lower bound at zero under the
smaller subset).}

\paragraph{Element textualisation stoplist.}
\rev{Queries and candidates use lowercased alphanumeric tokens with a
short UI stoplist, so lexical baselines are not boosted by generic UI
verbs or class names that appear in both query and candidate by
construction.}

\section{LCC/ESH thresholds and rank-caching requirements}
\label{app:lcc-thresholds}

We adopt three reading thresholds, calibrated against the cross-corpus
patterns in \S\ref{sec:results}. We use these thresholds as
descriptive reading aids, not as hypothesis tests:
\begin{itemize}\itemsep0pt\parsep0pt
    \item $\textrm{LCC}(M) > 0.85$: hit@1 outcomes are discriminated
        from misses with AUC above $0.85$ by the lexical and structural
        controls, without reference to the encoder.
    \item $\textrm{ESH}^{\textrm{oracle}}(\mathcal{S})
        - \textrm{ESH}^{\textrm{real}}(\mathcal{S}) \geq 0.05$ at R@1:
        in-principle complementary signal is largely unrealisable by
        standard RRF: the embeddings carry information BM25
        lacks but RRF cannot cheaply extract it.
    \item $\textrm{ESH}^{\textrm{real}}(\mathcal{S}) \leq 0$ at R@1:
        destructive fusion; adding $\mathcal{S}$ to BM25 under
        RRF hurts top-1 accuracy.
\end{itemize}

LCC needs only the per-step target rank under each retriever.
$\textrm{ESH}^{\textrm{real}}$ additionally requires per-candidate
per-method ranks at compute time (needed to compute each
candidate's RRF score), though only the target's combined rank is
retained afterwards. Both diagnostics extend to any GUI grounding
benchmark with element-level ground truth, a candidate pool, and at
least one lexical baseline.

\section{Statistical inference and implementation details}
\label{app:impl}

\paragraph{Models, hyperparameters, and compute.}
The encoders are off-the-shelf checkpoints used without modification:
MiniLM (\texttt{all-MiniLM-L6-v2}, 22M), MPNet
(\texttt{all-mpnet-base-v2}, 110M), BGE
(\texttt{BAAI/bge-large-en-v1.5}, 335M), E5
(\texttt{intfloat/e5-large-v2}, 335M), and \rev{Qwen3-Embedding-0.6B
(\texttt{Qwen/Qwen3-Embedding-0.6B}, 0.6B).} BM25 uses standard, untuned
parameters $k_1{=}1.5$, $b{=}0.75$, and reciprocal-rank fusion uses
the standard constant $k{=}60$ (not tuned). We perform no
\rev{hyperparameter search for the zero-shot retrievers. For Qwen3, queries
use the model-card template \texttt{Instruct: Given a GUI instruction,
retrieve the target UI element.} followed by \texttt{Query: \{instruction\}};
candidates use the same
UI-element textualisation as all other retrievers and no candidate-side
instruction. We use the checkpoint's default tokenizer truncation and
\texttt{SentenceTransformer.encode(..., normalize\_embeddings=True)}, then
score candidates by dot product over the normalised vectors.} The single
training run is a one-epoch contrastive fine-tune of MiniLM-L6-v2 with
\texttt{MultipleNegativesRankingLoss}, batch size $64$, learning rate
$2\mathrm{e}{-}5$, $10\%$ warmup, and up to three same-screen hard
negatives per example. All experiments run on a single consumer-grade
GPU with FP16 inference; the workload is dominated by embedding
extraction and per-screen candidate ranking (inference only), with the
one-epoch fine-tune as the sole training cost, for a total on the
order of a few GPU-hours. Encoders are loaded and fine-tuned with the
\texttt{sentence-transformers} library; Mind2Web DOM parsing uses
\texttt{BeautifulSoup}; bootstrap inference and correlations use
\texttt{NumPy}/\texttt{SciPy}; BM25 is our own implementation with the
parameters above. A pinned dependency list ships with the released
code.

All paired comparisons (single-retriever deltas, fusion deltas
versus BM25, and partial-correlation deltas) use a paired cluster
bootstrap with $B=2000$ resamples by \texttt{episode\_id} on
AndroidControl and MoTIF, and by \texttt{annotation\_id} on Mind2Web
(matching the corpus's unit of within-cluster correlation). Reporting
by episode/annotation respects the non-independence between
consecutive steps of the same task. 95\% confidence intervals are reported as
percentile intervals; we say an effect ``actively excludes zero''
when both endpoints lie on the same side of zero. Per-step Spearman
and Pearson correlations use the standard formulae without further
adjustment; we never report a $p$-value without the corresponding
effect-size CI.

All ranking ties are broken with a hash-based
deterministic jitter keyed on the episode id, step index, retriever,
and candidate index so that rank assignment is independent of
traversal order and reproducible across subsets. Random seeds, the
cluster-bootstrap RNG seed, and the BM25 hyperparameters are fixed;
the full pipeline is released alongside the per-step CSV panels
(license and artifact terms in Appendix~\ref{app:artifacts}).

\paragraph{BM25 IDF on small per-screen collections.}
Each screen is treated as its own BM25 collection, so IDF is estimated
from a median of 62 candidates and is necessarily coarse. This does not
drive the lexical advantage. In the limit of uninformative IDF, BM25
reduces to length-normalised query--candidate token overlap, the
same surface-label signal Jaccard captures. Jaccard tracks BM25
to within a few points of $R@1$ on every corpus
(Tables~\ref{tab:q7-main},~\ref{tab:replication}). The lexical lead
therefore reflects instruction--label overlap rather than precise IDF
calibration.

\section{Paired BM25-vs-encoder deltas on AndroidControl}
\label{app:deltas}

Table~\ref{tab:q7-delta} reports the paired cluster-bootstrap deltas
of BM25 over each encoder on clean-main, summarised in
\S\ref{sec:res-ranking}.

\begin{table}[h]
\centering\scriptsize
\setlength{\tabcolsep}{3pt}
\begin{tabular}{@{}lcc@{}}
\toprule
Encoder & $\Delta R@1$ [95\% CI] & $\Delta$MRR [95\% CI] \\
\midrule
MiniLM    & $+0.068$ [$+0.063,+0.073$] & $+0.040$ [$+0.037,+0.044$] \\
MPNet     & $+0.090$ [$+0.085,+0.096$] & $+0.049$ [$+0.044,+0.053$] \\
BGE-large & $+0.037$ [$+0.033,+0.042$] & $+0.016$ [$+0.012,+0.019$] \\
\rev{E5-large} & \rev{$+0.050$ [$+0.045,+0.055$]} & \rev{$+0.029$ [$+0.025,+0.032$]} \\
\rev{Qwen3-Emb.} & \rev{$+0.076$ [$+0.071,+0.082$]} & \rev{$+0.055$ [$+0.051,+0.059$]} \\
\bottomrule
\end{tabular}
\caption{Paired cluster-bootstrap deltas of BM25 over each encoder on
clean-main ($n{=}32{,}646$; $B{=}2000$ by \texttt{episode\_id}).
Positive favours BM25; all CIs exclude zero.}
\label{tab:q7-delta}
\end{table}

\section{Label-type collapse heatmap}
\label{app:label-collapse}

\begin{figure*}[h]
\centering
\includegraphics[width=0.85\linewidth]{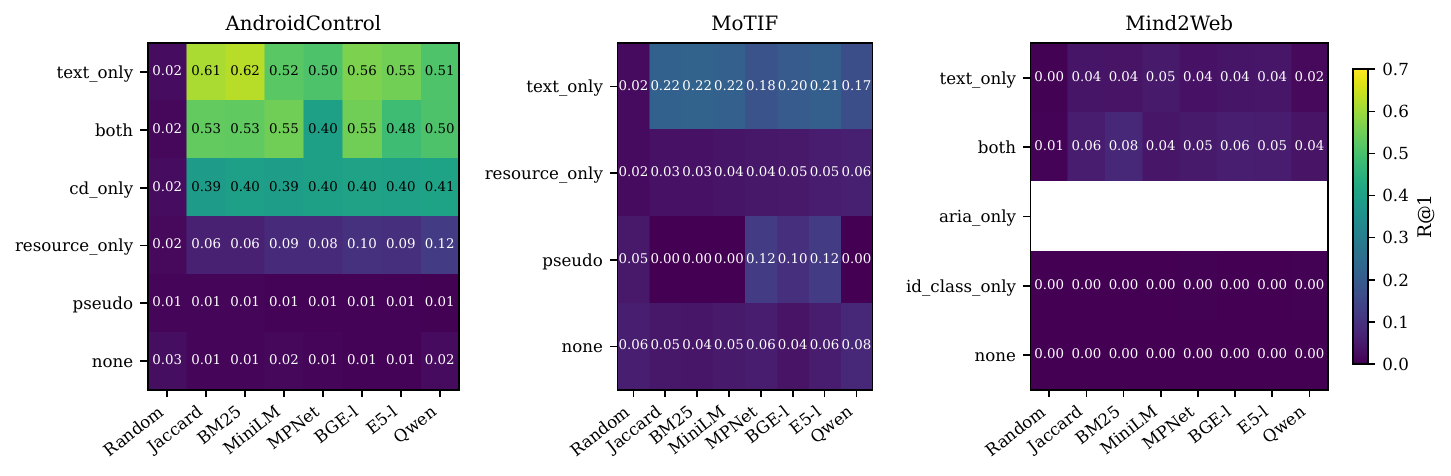}
\caption{Per-cell $R@1$ on AndroidControl (all DB rows,
$n{=}58{,}078$; six-way label-type partition), MoTIF (all steps,
$n{=}22{,}427$; four-way partition: text-only, resource-only,
pseudo, none), and Mind2Web (all test splits, $n{=}5{,}941$;
five-way label-type partition).}
\label{fig:label-collapse}
\end{figure*}

\section{Target-text dependence perturbation}
\label{app:label-mask}

To quantify how strongly target ranking depends on text exposed by the
gold candidate, we run an asymmetric target-label ablation on
AndroidControl clean-main. We use equal allocation across three label
strata, sampling $1{,}000$ \textsc{text\_only}, $1{,}000$
\textsc{cd\_only}, and $1{,}000$ \textsc{both} rows. For each row, we
blank only the target candidate's \texttt{text} and
\texttt{content\_description}; the query, candidate pool,
non-target candidates, target \texttt{resource\_id}, and target class
name remain fixed. The target textualisation changes for
$2{,}998/3{,}000$ rows. The resulting $R@1$ drops are large for BM25
and all five single-vector encoders.
The delta measures dependence on target-text availability. Because a UI
label carries both lexical and semantic information, the perturbation is
most informative as part of the full audit suite, alongside the lexical
baselines, label-type stratification, and LCC.
After masking, candidate textualisation falls back to the target's
\texttt{resource\_id} or class name. Resource IDs can retain label-related
tokens, so the perturbation conservatively removes the human-facing fields
rather than every potentially informative target string. The aggregate
gives equal weight to the three sampled label strata and is a diagnostic
average rather than a corpus-prevalence-weighted estimate.

\begin{table}[h]
\centering\scriptsize
\setlength{\tabcolsep}{3pt}
\begin{tabular}{@{}lccc@{}}
\toprule
Method & Orig. & Masked & $\Delta$ [95\% CI] \\
\midrule
BM25      & 0.508 & 0.125 & $-0.383$ [$-0.402,-0.364$] \\
MiniLM    & 0.487 & 0.165 & $-0.321$ [$-0.343,-0.300$] \\
MPNet     & 0.434 & 0.139 & $-0.296$ [$-0.316,-0.275$] \\
BGE-large & 0.503 & 0.188 & $-0.315$ [$-0.336,-0.294$] \\
E5-large  & 0.470 & 0.159 & $-0.311$ [$-0.330,-0.291$] \\
\rev{Qwen3-Emb.} & \rev{0.466} & \rev{0.251} & \rev{$-0.216$ [$-0.237,-0.196$]} \\
\bottomrule
\end{tabular}
\caption{Target-text dependence on an equal-allocation stratified
AndroidControl clean-main sample ($n{=}3{,}000$; $1{,}000$ per label
stratum). Negative deltas measure the loss in top-1 recovery when the gold
candidate's exposed label is removed while the query and non-target
candidates stay fixed.}
\label{tab:label-mask}
\end{table}

\section{LCC cross-validated AUC}
\label{app:lcc-cv}

We report five-fold grouped cross-validated AUC of the LCC logistic
regression alongside the in-sample AUC used in
Table~\ref{tab:cross-corpus}. Folds are split by \texttt{episode\_id}
on AndroidControl and MoTIF, and by \texttt{annotation\_id} on
Mind2Web (matching the cluster unit used elsewhere for these
corpora), so that no episode / annotation appears in both train and
test of any fold.

\begin{table}[h]
\centering\small
\begin{tabular}{l l cc}
\toprule
Set & Enc. & Full AUC & CV AUC \\
\midrule
AC main            & MiniLM     & 0.855 & 0.855 \\
                   & MPNet      & 0.834 & 0.834 \\
                   & BGE-large  & 0.880 & 0.880 \\
                   & \rev{E5-large} & \rev{0.882} & \rev{0.882} \\
                   & \rev{Qwen3-Emb.} & \rev{0.860} & \rev{0.860} \\
\midrule
MoTIF              & MiniLM     & 0.868 & 0.867 \\
                   & MPNet      & 0.825 & 0.821 \\
                   & BGE-large  & 0.845 & 0.844 \\
                   & \rev{E5-large} & \rev{0.874} & \rev{0.874} \\
                   & \rev{Qwen3-Emb.} & \rev{0.851} & \rev{0.849} \\
\midrule
M2W all            & MiniLM     & 0.849 & 0.846 \\
                   & MPNet      & 0.844 & 0.840 \\
                   & BGE-large  & 0.867 & 0.864 \\
                   & \rev{E5-large} & \rev{0.861} & \rev{0.855} \\
                   & \rev{Qwen3-Emb.} & \rev{0.845} & \rev{0.832} \\
\midrule
M2W rich           & MiniLM     & 0.836 & 0.837 \\
                   & MPNet      & 0.839 & 0.839 \\
                   & BGE-large  & 0.855 & 0.854 \\
                   & \rev{E5-large} & \rev{0.849} & \rev{0.849} \\
                   & \rev{Qwen3-Emb.} & \rev{0.835} & \rev{0.825} \\
\bottomrule
\end{tabular}
\caption{LCC full-data AUC vs.\ 5-fold episode-grouped CV AUC. The two
diverge by at most $0.013$ (Qwen3-Embedding-0.6B, M2W all), confirming that the
lexical coupling signal is not an in-sample artefact.}
\label{tab:lcc-cv}
\end{table}

\section{Oracle min-rank vs.\ true RRF: full table}
\label{app:fusion}

\begin{figure}[!htbp]
\centering
\includegraphics[width=\linewidth]{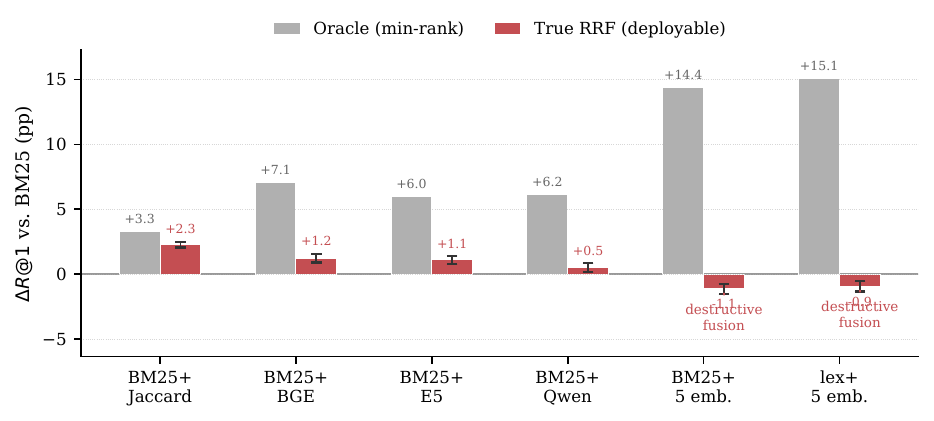}
\caption{Oracle min-rank ($\textrm{ESH}^{\textrm{oracle}}$, grey) vs.\
deployable reciprocal-rank fusion ($\textrm{ESH}^{\textrm{real}}$, red)
$\Delta R@1$ over BM25 on AndroidControl clean-main. Error bars show
$B{=}2000$ cluster-bootstrap $95\%$ CIs for deployable RRF deltas;
oracle bars are target-aware upper bounds.}
\label{fig:oracle-vs-rrf}
\end{figure}

Table~\ref{tab:fusion} gives the full candidate-level oracle-vs-RRF comparison
summarised in \S\ref{sec:res-residual} and plotted in
Figure~\ref{fig:oracle-vs-rrf}. Oracle min-rank uses the privileged
target rank per row to upper-bound any combiner; true RRF combines
per-candidate ranks and is therefore deployable. The two diverge
sharply: oracle lifts $R@1$ by up to \rev{$+0.151$}, while the same
combiner under true RRF \emph{loses} $R@1$. \rev{Single-embedding RRF
pairs with BGE, E5, and Qwen3 give only small deployable gains, and the
largest deployable top-1 lift comes from the lexical pair
BM25\,$+$\,Jaccard.}

\begin{table*}[!htbp]
\centering\small
\begin{tabular}{l cc cc l}
\toprule
& \multicolumn{2}{c}{Oracle min-rank} & \multicolumn{2}{c}{True RRF} & \\
\cmidrule(lr){2-3}\cmidrule(lr){4-5}
Combiner & $R@1$ & $\Delta$ vs BM25 & $R@1$ & $\Delta$ vs BM25 & MRR (RRF) \\
\midrule
BM25                      & 0.555 & -- & 0.555 & -- & 0.631 \\
BM25 $+$ Jaccard          & 0.588 & $+0.033$ & \textbf{0.578} & $+0.023$~[$+0.021,+0.025$] & \textbf{0.642} \\
BM25 $+$ BGE              & 0.626 & $+0.071$ & 0.568 & $+0.012$~[$+0.009,+0.016$] & 0.640 \\
\rev{BM25 $+$ E5}         & \rev{0.615} & \rev{$+0.060$} & \rev{0.566} & \rev{$+0.011$~[$+0.008,+0.014$]} & \rev{0.638} \\
\rev{BM25 $+$ Qwen3}      & \rev{0.617} & \rev{$+0.062$} & \rev{0.560} & \rev{$+0.005$~[$+0.002,+0.008$]} & \rev{0.630} \\
\rev{BM25 $+$ 5 emb.}     & \rev{0.699} & \rev{$+0.144$} & \rev{0.544} & \rev{$-0.011$~[$-0.016,-0.007$]} & \rev{0.638} \\
\rev{\textbf{lex $+$ 5 emb.}} & \rev{\textbf{0.706}} & \rev{\textbf{$+0.151$}} & \rev{0.546} & \rev{$-0.009$~[$-0.013,-0.005$]} & \rev{0.641} \\
\bottomrule
\end{tabular}
\caption{Oracle min-rank vs.\ true RRF on AndroidControl clean-main
($n{=}32{,}646$). True-RRF $\Delta$ vs BM25 95\% CIs are paired
cluster bootstrap by \texttt{episode\_id} with $B{=}2000$. \rev{``5 emb.''
denotes MiniLM, MPNet, BGE-large, E5-large, and Qwen3-Embedding-0.6B;
``lex'' adds Jaccard to BM25.}}
\label{tab:fusion}
\end{table*}

\section{Validation-tuned calibrated fusion stress test}
\label{app:calibrated-fusion}

\paragraph{\rev{Calibrated fusion stress test.}}
\rev{Beyond the fixed RRF rule used in the main analysis, we test
whether a minimal validation-tuned linear score fusion recovers more
of the oracle headroom on AndroidControl clean-main. For each screen,
raw BM25 scores and embedding cosine scores are z-normalized over the
candidate pool, then combined as
$s_{\alpha}(c)=\alpha z_{\mathrm{BM25}}(c)+(1-\alpha)z_{\mathrm{emb}}(c)$.
A single interpolation weight $\alpha\in\{0,0.05,\ldots,1\}$ is selected on
the official validation split ($n{=}261$ clean-main rows) to maximize
$R@1$, with MRR as a tie-breaker, and then evaluated on the held-out test
split ($n{=}2{,}949$). This is not a learned reranker, but a simple
calibration check.}

\begin{table*}[!htbp]
\centering\small
\begingroup
\setlength{\tabcolsep}{3pt}
\begin{tabular}{l l c c c c c}
\toprule
Method & Tuning & $R@1$ & $R@5$ & $R@10$ & MRR & $\Delta R@1$ vs BM25 (pp) \\
\midrule
BM25 & -- & 0.558 & 0.694 & 0.733 & 0.628 & -- \\
MiniLM & -- & 0.518 & 0.717 & 0.775 & 0.611 & $-3.93$ [$-5.60,-2.44$] \\
BM25+MiniLM RRF & fixed $k{=}60$ & 0.567 & 0.702 & 0.758 & 0.635 & $+0.95$ [$-0.03,+1.94$] \\
BM25+MiniLM calibrated & val $\alpha{=}0.60$ & 0.580 & 0.747 & 0.794 & 0.659 & $+2.27$ [$+1.35,+3.23$] \\
Oracle BM25+MiniLM upper & target-aware & 0.625 & 0.779 & 0.825 & 0.696 & $+6.71$ [$+5.76,+7.65$] \\
MPNet & -- & 0.492 & 0.729 & 0.789 & 0.603 & $-6.54$ [$-8.29,-4.80$] \\
BM25+MPNet RRF & fixed $k{=}60$ & 0.569 & 0.716 & 0.777 & 0.642 & $+1.19$ [$+0.16,+2.19$] \\
BM25+MPNet calibrated & val $\alpha{=}0.65$ & 0.586 & 0.754 & 0.803 & 0.668 & $+2.81$ [$+1.89,+3.81$] \\
Oracle BM25+MPNet upper & target-aware & 0.625 & 0.786 & 0.838 & 0.702 & $+6.78$ [$+5.83,+7.76$] \\
BGE-large & -- & 0.513 & 0.721 & 0.790 & 0.613 & $-4.41$ [$-5.97,-2.83$] \\
E5-large & -- & 0.516 & 0.719 & 0.773 & 0.610 & $-4.10$ [$-5.63,-2.57$] \\
Qwen3-Emb. & -- & 0.486 & 0.680 & 0.743 & 0.579 & $-7.19$ [$-8.72,-5.55$] \\
BM25+BGE RRF & fixed $k{=}60$ & 0.562 & 0.697 & 0.766 & 0.633 & $+0.44$ [$-0.60,+1.52$] \\
BM25+BGE calibrated & val $\alpha{=}1.00$ & 0.558 & 0.694 & 0.733 & 0.628 & $0.00$ [$0.00,0.00$] \\
Oracle BM25+BGE upper & target-aware & 0.626 & 0.791 & 0.848 & 0.706 & $+6.88$ [$+5.95,+7.82$] \\
BM25+E5 RRF & fixed $k{=}60$ & 0.564 & 0.699 & 0.761 & 0.633 & $+0.64$ [$-0.33,+1.63$] \\
BM25+E5 calibrated & val $\alpha{=}0.75$ & 0.572 & 0.748 & 0.796 & 0.656 & $+1.46$ [$+0.52,+2.42$] \\
Oracle BM25+E5 upper & target-aware & 0.617 & 0.779 & 0.827 & 0.693 & $+5.97$ [$+5.08,+6.88$] \\
BM25+Qwen3 RRF & fixed $k{=}60$ & 0.552 & 0.687 & 0.756 & 0.624 & $-0.54$ [$-1.63,+0.60$] \\
BM25+Qwen3 calibrated & val $\alpha{=}0.90$ & 0.575 & 0.736 & 0.779 & 0.654 & $+1.80$ [$+0.87,+2.79$] \\
Oracle BM25+Qwen3 upper & target-aware & 0.619 & 0.763 & 0.816 & 0.690 & $+6.17$ [$+5.25,+7.09$] \\
\bottomrule
\end{tabular}
\caption{Validation-tuned calibrated score fusion on AndroidControl clean-main
test ($n{=}2{,}949$). Scores are z-normalized within each candidate pool;
$\alpha$ is selected on validation by $R@1$ with MRR as a tie-breaker.
Oracle rows are target-aware upper bounds, not deployable systems. CIs use
paired cluster bootstrap by episode with $B{=}2000$.}
\label{tab:calibrated-fusion}
\endgroup
\end{table*}

\rev{Calibrated fusion recovers limited deployable signal across the five
main encoders. Validation-tuned fusion yields statistically stable gains for
MiniLM ($+2.27$~pp, CI $[+1.35,+3.23]$), MPNet ($+2.81$~pp, CI
$[+1.89,+3.81]$), E5 ($+1.46$~pp, CI $[+0.52,+2.42]$), and Qwen3
($+1.80$~pp, CI $[+0.87,+2.79]$). For BGE, validation selects
$\alpha{=}1.00$, so the calibrated system reduces exactly to BM25 and yields
no held-out change. The $n{=}261$ validation subset limits the reliability of
$\alpha$ selection; this outcome should be read as ``no consistent gain
detected on validation'', not necessarily ``no signal exists''.
All five calibrated results remain far below their
corresponding oracle headroom ($+5.97$ to $+6.88$~pp). Fixed RRF yields only
small held-out changes ($-0.54$ to $+1.19$~pp), with a significant gain only
for BM25+MPNet. Thus, calibrated fusion supports
the same measurement recommendation: oracle headroom and deployable fusion
gains should be reported separately.}

\FloatBarrier
\section{Fine-tuning: full breakdown and coupling audit}
\label{app:finetune}

We fine-tune MiniLM-L6-v2 with a one-epoch contrastive objective on
the AndroidControl training split. We sample $29{,}426$ training
examples from $29{,}436$ clean-main train steps; each example carries
a query, the target candidate, and up to three same-screen hard
negatives, in addition to the in-batch negatives provided by
MultipleNegativesRankingLoss. The evaluation protocol mirrors the
zero-shot evaluation: per-screen candidate ranking on the held-out
$2{,}949$-row clean-main test subset.

\begin{table}[h]
\centering\small
\setlength{\tabcolsep}{4pt}
\begin{tabular}{l r cccc}
\toprule
Split & $n$ & BM25 & $M_0$ & \textbf{$M_{\mathrm{ft}}$} & BGE$_0$ \\
\midrule
Overall                & 2{,}949 & 0.558 & 0.518 & \textbf{0.694} & 0.513 \\
IDD                    & 1{,}536 & 0.548 & 0.490 & \textbf{0.703} & 0.501 \\
\texttt{app\_unseen}   &     36  & 0.778 & 0.722 & \textbf{0.833} & 0.694 \\
\texttt{task\_unseen}  & 1{,}377 & 0.562 & 0.544 & \textbf{0.680} & 0.523 \\
\midrule
text-only    & 1{,}784 & 0.607 & 0.523 & \textbf{0.678} & 0.534 \\
cd-only      &    903  & 0.458 & 0.485 & \textbf{0.711} & 0.452 \\
both          &    262  & 0.561 & 0.595 & \textbf{0.737} & 0.584 \\
\bottomrule
\end{tabular}
\caption{Full $R@1$ breakdown of fine-tuning evaluation on the
AndroidControl test clean-main subset.
$M_0$ = zero-shot MiniLM-L6-v2;
$M_{\mathrm{ft}}$ = same model after one epoch;
BGE$_0$ = zero-shot BGE-large.}
\label{tab:finetune-full}
\end{table}

\rev{We rerun two complementary diagnostics on the same held-out test
rows (Tables~\ref{tab:finetune-coupling}--\ref{tab:finetune-mask}): LCC
measures whether supervision changes lexical coupling, while the target-text
perturbation measures whether the improved ranker remains dependent on text
exposed by the gold candidate. Fine-tuning substantially improves target
recovery, including on BM25 misses, but the hit pattern remains highly
predictable from lexical and structural controls, and removing the gold
candidate's exposed text sharply reduces top-1 accuracy.}

\begin{table}[h]
\centering\small
\setlength{\tabcolsep}{4pt}
\begin{tabular}{lccc}
\toprule
Method & $R@1$ & LCC & BM25-miss rec. \\
\midrule
$M_0$ & 0.518 & 0.865 & 0.151 \\
$M_{\mathrm{ft}}$ & \textbf{0.694} & 0.877 & \textbf{0.374} \\
BGE$_0$ & 0.513 & 0.868 & 0.156 \\
\bottomrule
\end{tabular}
\caption{\rev{Fine-tuned MiniLM coupling audit on the same held-out
AndroidControl clean-main test rows ($n{=}2{,}949$). LCC is 5-fold
episode-grouped CV AUC using BM25 rank, Jaccard rank, candidate-pool
size, and label type. BM25-miss recovery is the fraction of rows with
BM25 rank $>1$ that the method ranks first.}}
\label{tab:finetune-coupling}
\end{table}

\begin{table}[h]
\centering\small
\setlength{\tabcolsep}{4pt}
\begin{tabular}{lccc}
\toprule
Method & Orig. & Masked & $\Delta$ [95\% CI] \\
\midrule
BM25 & 0.558 & 0.134 & $-0.423$ [$-0.444,-0.402$] \\
$M_0$ & 0.518 & 0.178 & $-0.340$ [$-0.363,-0.316$] \\
$M_{\mathrm{ft}}$ & 0.694 & 0.151 & $-0.542$ [$-0.562,-0.520$] \\
\bottomrule
\end{tabular}
\caption{\rev{Held-out target-text dependence perturbation for the
fine-tuning control ($R@1$). The perturbation blanks the gold candidate's
visible text and content description while retaining the query, non-target
candidates, resource IDs, class names, and candidate pools. Negative deltas
measure sensitivity to target-exposed text.}}
\label{tab:finetune-mask}
\end{table}

\FloatBarrier
\section{Construct correlation: M3 converges to Jaccard}
\label{app:construct}

Throughout this appendix and Appendix~\ref{app:validity} we write
$M_3(q, t)$ for the per-step instruction--target cosine distance
$1 - \mathrm{cos}(\phi(q), \phi(t))$ under a sentence encoder $\phi$,
and $M_1$ for its episode-level analogue (defined in
Appendix~\ref{app:validity}). $M_3$ is the step-level alignment
quantity that the Limitations section refers to when bounding our
claim to grounding-metric validity rather than behavioural
prediction.

Beyond ranking, we measure how strongly each embedding family's
step-level distance to the textualised target correlates with a
matching Jaccard baseline. Three textualisation variants are compared:
current (the form used for ranking), label-only
(text + content description, no class or action prefix), and
action-aware (action verb prepended). For every variant and
embedding family we compute the per-step cosine distance and the
per-step Jaccard distance between the same query and the same
textualised target, then report Spearman correlation across the
clean-strict subset of the step-local panel ($n=32{,}636$ steps with
a valid M3, target text or content description, and at least one
subsequent action).

On this subset, M3 is highly correlated with the matching token-level
Jaccard distance across all three textualisation variants and all five
encoder families
(Table~\ref{tab:m3-jaccard}); Spearman clusters
in $[0.68,\, 0.82]$, with action-aware textualisation producing the
tightest coupling for MiniLM, MPNet, and BGE-large (E5-large is the
exception, where the \emph{current} textualisation is marginally
tighter than action-aware). Model-to-model agreement (same variant) is
even higher, with pairwise Spearman in $[0.80,\, 0.91]$.

\begin{table}[h]
\centering\small
\renewcommand{\arraystretch}{1.2}
\setlength{\tabcolsep}{7pt}
\begin{tabular}{l ccc}
\toprule
Model & current & label-only & action-aware \\
\midrule
MiniLM     & +0.750 & +0.767 & +0.800 \\
MPNet      & +0.684 & +0.704 & +0.718 \\
BGE-large  & +0.749 & +0.772 & +0.803 \\
E5-large   & +0.821 & +0.798 & +0.814 \\
Qwen3-Emb. & +0.714 & +0.727 & +0.797 \\
\bottomrule
\end{tabular}
\caption{Step-level M3--Jaccard Spearman ($n{=}32{,}636$).
All cells $p < 10^{-300}$.}
\label{tab:m3-jaccard}
\end{table}

\FloatBarrier
\section{STS-B per-encoder Spearman}
\label{app:sts}

\begin{center}
\centering\small
\setlength{\tabcolsep}{4pt}
\begin{tabular}{l cc}
\toprule
Retriever & Spearman vs.\ gold & Spearman vs.\ Jaccard \\
\midrule
Jaccard   & $+0.651$ & -- \\
MiniLM    & $+0.845$ & $+0.687$ \\
MPNet     & $+0.859$ & $+0.647$ \\
BGE-large & $+0.886$ & $+0.749$ \\
E5-large  & $+0.886$ & $+0.733$ \\
Qwen3-Emb. & $+0.814$ & $+0.608$ \\
\bottomrule
\end{tabular}
\captionof{table}{Spearman correlations on STS-B dev+test ($2{,}879$ pairs). All five
encoders beat Jaccard against gold similarity by $16$--$24$
points; embedding--Jaccard agreement remains in
$[0.608, 0.749]$, lower than the corresponding $[0.68, 0.82]$ range
on AndroidControl clean-strict (Appendix~\ref{app:construct}).}
\label{tab:sts-full}
\end{center}
\FloatBarrier

\section{Behavioural validity boundary}
\label{app:validity}

We test whether instruction-target alignment metrics predict
behavioural friction proxies at both the step and the episode level.

\paragraph{Step-level friction labels.}
\rev{For each step in the no-duplicate subset, we mark whether either
of the next two actions is a back, wait, or scroll action. We also
use two robustness variants: a non-terminal back-out label that
excludes repeated terminal exits, and a broader friction label that
also counts home/open-app transitions. For each alignment metric and
friction proxy we report marginal Spearman, decile contrasts, and
within-episode pairwise deltas comparing the highest- and lowest-M3
steps in the same episode.}

\rev{Step-level M3 variants do not predict local repair signals. On
the no-duplicate subset ($n{=}26{,}832$), current M3 is near zero
against back actions ($-0.008$~[$-0.020,+0.004$]), non-terminal
back-outs ($-0.007$~[$-0.018,+0.005$]), and the combined
back/wait/scroll friction proxy ($-0.009$~[$-0.020,+0.003$]). The
action-aware variant is similar in magnitude and direction. A lexical
Jaccard baseline on the same labels is slightly larger but
same-direction (combined-friction Spearman
$-0.020$~[$-0.031,-0.008$]), indicating that the step-local null is
not an embedding-specific artefact.}

\paragraph{Episode-level mismatch.}
Let $g$ be an episode's goal string and $L = (l_1,\dots,l_n)$ its
prospective step instructions. The full-TFRecord episode-level
mismatch is
\begin{equation}
M_1(g,L) = 1 - \mathrm{cos}\bigl(\phi(g),\,\phi(l_1 \,\Vert\,
\cdots \,\Vert\, l_n)\bigr),
\end{equation}
where $\phi$ is the MiniLM-L6 encoder and $\Vert$ denotes
whitespace-joined concatenation. We compute $M_1$ over all 15{,}283
episodes and \rev{correlate it against episode-level back, wait, and
scroll frequencies and incidence indicators using marginal Spearman
and partial Spearman controlling for episode length, instruction
diversity, and goal--step lexical distance.}

\rev{Full $M_1$ has small but bounded positive associations with
aggregate friction: marginal Spearman is $+0.092$~[$+0.077,+0.107$]
for back frequency and $+0.124$~[$+0.108,+0.140$] for scroll
frequency. After the controls, these shrink to
$+0.102$~[$+0.088,+0.117$] and $+0.038$~[$+0.023,+0.053$], below
0.15 throughout. We read this as weak task-structure signal rather
than direct semantic difficulty: full $M_1$ correlates with
goal--step Jaccard distance at Spearman $+0.715$, and removing the
lexical and length components shrinks the residual to small positive
values.}

\paragraph{Typing-correction sparsity.}
Typing-correction events (consecutive same-field re-entries)
occur in only 82 instances across 63 of the 15{,}283 episodes
($0.4\%$ episode incidence), too sparse to serve as a friction proxy.
This confirms the broader pattern: AndroidControl annotators were
instructed to avoid unnecessary or unrelated actions, so behavioural
friction surfaces are systematically sparse.

\section{Qualitative error analysis}
\label{app:diagnostic-examples}

\rev{Table~\ref{tab:diagnostic-examples} reports de-identified
problematic examples from all three datasets. Rows are selected to
illustrate three recurring regimes in the observed ranks: lexical
label recovery, residual embedding recovery, and label-poor failure.
Text fields are paraphrased to avoid exposing user or third-party
dataset content, while ranks and label types are unchanged. For compactness,
rank summaries report BM25 and four representative encoders; all five
encoders are covered in the quantitative analyses.}

\begin{center}
\begingroup
\scriptsize
\setlength{\tabcolsep}{1.5pt}
\begin{tabular}{@{}p{0.10\linewidth} p{0.13\linewidth} p{0.48\linewidth} p{0.23\linewidth}@{}}
\toprule
Data & Regime & De-identified pattern and diagnosis & Label/ranks \\
\midrule
AC &
Lexical recovery &
Query names a visible \textsc{Delete} control in an alarm screen; the
target text is \textsc{Delete}. Visible label recovery is sufficient;
several encoders rank other alarm/time candidates higher. &
text-only; BM25 1; Mini 21; MPNet 21; BGE 11; E5 1 \\

AC &
Residual &
Query asks to open app preferences/options; the target label is
\textsc{Settings}. The embedding rankers recover a close lexical/semantic
variant that BM25 misses. &
text-only; BM25 145; Mini 1; MPNet 2; BGE 1; E5 1 \\

AC &
Label-poor &
Query chooses an overflow icon for a media item; the target exposes only
a generic resource id. The candidate text lacks the item title and
visible affordance, so all text rankers miss. &
resource-only; BM25 85; Mini 102; MPNet 54; BGE 61; E5 41 \\

MoTIF &
Lexical recovery &
In a mobile time/weather workflow, the target exposes visible text that
matches the local instruction. Lexical matching recovers the target, while
one embedding ranker is pulled to another time-related candidate. &
text-only; BM25 1; Mini 1; MPNet 17; BGE 1; E5 1 \\

MoTIF &
Residual &
In a mobile shopping workflow, the target is text-labelled but not a
BM25 hit. Dense encoders rank the target first, indicating residual
signal beyond exact label recovery in this row. &
text-only; BM25 38; Mini 1; MPNet 1; BGE 1; E5 1 \\

MoTIF &
Label-poor &
In a dense mobile screen, the target has no readable text. All text-only
rankers place the target near the bottom of a large candidate pool. &
none; BM25 150; Mini 166; MPNet 152; BGE 167; E5 132 \\

M2W &
Lexical recovery &
On an information website, a visible link/button string directly matches
the action. BM25 ranks the target first, while embeddings still move
nearby content labels above it. &
text-only; BM25 1; Mini 6; MPNet 5; BGE 12; E5 7 \\

M2W &
Residual &
On a service website, the target is text-labelled but BM25 ranks it
below many candidates. Dense encoders recover it near the top, consistent
with residual signal beyond exact lexical matching. &
text-only; BM25 65; Mini 4; MPNet 2; BGE 1; E5 3 \\

M2W &
Label-poor &
In a large DOM form, the target exposes only id/class metadata. Without
readable target text, all rankers miss top-1 despite some embedding
improvement over BM25. &
id/class-only; BM25 511; Mini 78; MPNet 296; BGE 209; E5 61 \\
\bottomrule
\end{tabular}
\captionof{table}{Qualitative error analysis for recurring ranking regimes.
AC abbreviates AndroidControl and M2W abbreviates Mind2Web. Text
is paraphrased to avoid exposing user or third-party dataset content; the
tabulated ranks and label types are unchanged from the source rows.
Lower rank is better; rank~1 indicates the
top-ranked target.}
\label{tab:diagnostic-examples}
\endgroup
\end{center}

\section{Additional embedding and retrieval stress tests}
\label{app:stress-tests}

\rev{Table~\ref{tab:stress-tests} reports the paired $R@1$ intervals and
coupling diagnostics for Qwen3-Embedding-0.6B, which is included in the main
single-vector encoder comparison, and for one appendix-only broader retrieval
stress test. SPLADE (\citealp{formal2021splade};
\texttt{naver/splade-cocondenser-ensembledistil}) is included separately as a
learned sparse retriever, not as a main single-vector encoder baseline. These
rows show that the measurement diagnosis persists for both a more recent
single-vector embedding control and a broader sparse retrieval architecture:
Qwen3 remains below BM25 at $R@1$ on AndroidControl clean-main, MoTIF
clean-main, and Mind2Web all test splits, while SPLADE is below BM25 on
AndroidControl clean-main and MoTIF clean-main and statistically tied with
BM25 on Mind2Web all test splits; both remain highly predictable from lexical
and candidate controls.}

\begin{table}[h]
\centering\scriptsize
\setlength{\tabcolsep}{2pt}
\begingroup
\resizebox{\linewidth}{!}{%
\begin{tabular}{l l r c c c c}
\toprule
Stress test & Corpus & $n$ & BM25 & Method & $\Delta$ pp & Coupling AUC \\
\midrule
Qwen3-Emb. & AC main & 32{,}646 & 0.557 & 0.481 & $-7.64$ [$-8.22,-7.10$] & 0.860 \\
Qwen3-Emb. & MoTIF & 4{,}724 & 0.224 & 0.167 & $-5.72$ [$-6.97,-4.44$] & 0.851 \\
Qwen3-Emb. & M2W all & 5{,}941 & 0.041 & 0.020 & $-2.04$ [$-2.62,-1.45$] & 0.845 \\
SPLADE & AC main & 32{,}646 & 0.557 & 0.535 & $-2.21$ [$-2.65,-1.77$] & 0.911 \\
SPLADE & MoTIF & 4{,}724 & 0.224 & 0.213 & $-1.12$ [$-2.22,-0.04$] & 0.893 \\
SPLADE & M2W all & 5{,}941 & 0.041 & 0.042 & $+0.12$ [$-0.45,+0.67$] & 0.888 \\
\bottomrule
\end{tabular}
}
\caption{Qwen3 paired intervals and sparse-retrieval stress test. Deltas are
method minus BM25 at $R@1$; negative deltas mean the method is below BM25.
For SPLADE, Coupling AUC is the same hit@1 predictability
diagnostic, not the encoder-specific LCC definition.}
\label{tab:stress-tests}
\endgroup
\end{table}

\end{document}